\documentclass[11pt]{article}

\usepackage[final]{acl}
\usepackage{booktabs, makecell}
\usepackage{enumitem}
\usepackage{amsmath, amssymb}
\usepackage{array}
\usepackage{booktabs}
\usepackage{multirow}
\usepackage{times}
\usepackage{latexsym}
\usepackage{microtype}
\usepackage{xcolor}
\usepackage{soul}
\usepackage[ruled,vlined]{algorithm2e}
\sethlcolor{yellow!25}
\newcommand{\kw}[1]{\hl{#1}}
\usepackage[T1]{fontenc}

\usepackage[utf8]{inputenc}

\usepackage{microtype}

\usepackage{inconsolata}

\usepackage{graphicx}
\usepackage[table]{xcolor}  
\usepackage{colortbl}       

\definecolor{em}{gray}{0.9}

\usepackage{colortbl}
\definecolor{bestbg}{RGB}
{255,235,235}   
\definecolor{secondbg}{RGB}{235,245,255} 
\usepackage{tabularx,booktabs}
\newcolumntype{L}[1]{>{\raggedright\arraybackslash}p{#1}}
\newcolumntype{Y}{>{\raggedright\arraybackslash}X}
\usepackage{xcolor}
\usepackage{soul}
\definecolor{lightpink}{HTML}{F7DADA}

\title{FinAnchor: Aligned Multi-Model Representations for Financial Prediction}

\author{
\textbf{Zirui He\textsuperscript{1,*}} \,
\textbf{Huopu Zhang\textsuperscript{2,*}} \,
Yanguang Liu\textsuperscript{1} \,
Sirui Wu\textsuperscript{3} \,
Mengnan Du\textsuperscript{3}\\
\textsuperscript{1}New Jersey Institute of Technology \,
\textsuperscript{2}Georgia Institute of Technology \, \\
\textsuperscript{3}Chinese University of Hong Kong, Shenzhen\\
\small\texttt{\{yanguang.liu, zh296\}@njit.edu}, 
\small\texttt{hzhang931@gatech.edu},\\ 
\small\texttt{123090629@link.cuhk.edu.cn}, 
\small\texttt{mengnandu@cuhk.edu.cn}\\[2pt]
\small\textsuperscript{*}Co--first authors
}

\begin{document}
\maketitle
\begin{abstract}
Financial prediction from long documents involves significant challenges, as actionable signals are often sparse and obscured by noise, and the optimal LLM for generating embeddings varies across tasks and time periods. In this paper, we propose \textbf{FinAnchor} (Financial Anchored Representations), a lightweight framework that integrates embeddings from multiple LLMs without fine-tuning the underlying models. FinAnchor addresses the incompatibility of feature spaces by selecting an anchor embedding space and learning linear mappings to align representations from other models into this anchor. These aligned features are then aggregated to form a unified representation for downstream prediction. Across multiple financial NLP tasks, FinAnchor consistently outperforms strong single-model baselines and standard ensemble methods, demonstrating the effectiveness of anchoring heterogeneous representations for robust financial prediction.
\end{abstract}


\section{Introduction}

Financial text prediction, which aims to infer market-relevant outcomes from documents such as earnings-call transcripts, regulatory disclosures, and policy communications, has become a central problem at the intersection of finance and natural language processing ~\cite{skinner2002earnings,si2013exploiting,shah-etal-2023-trillion}. These tasks typically involve long-form documents: they are lengthy, highly structured, and noisy, containing substantial boilerplate and recurring discussion patterns, while truly decision-relevant signals are often sparse and subtle, frequently manifested as small shifts in guidance language, risk phrasing, or indications of operational pressure~\cite{brown2004conference,loughran2011liability,bushee2018linguistic,zaheer2020big}.

\begin{figure*}[t]
  \centering
  \includegraphics[width=\textwidth]{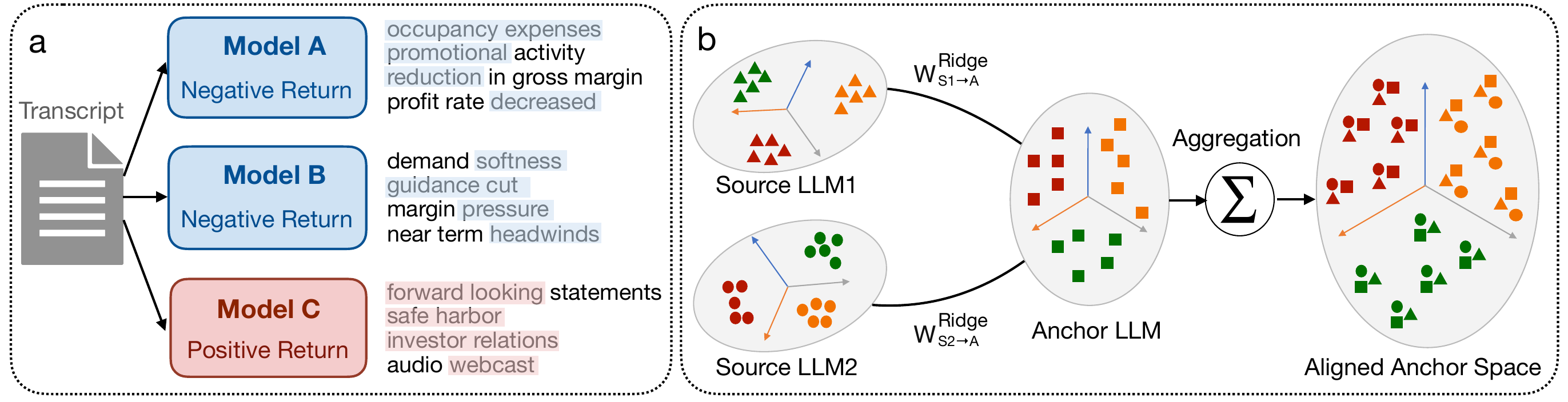}
  \caption{
  \textbf{(a)} Three different LLM independently predict next-day stock movement from the same earnings call transcript: Models A and B predict negative return, while Model C predicts positive return.
  Highlighted phrases denote text snippets deemed salient by each model. In this case, Model A and B emphasize financially relevant cues (e.g., margins, demand, guidance), while Model C is dominated by less relevant content.
  \textbf{(b)} \emph{Multi-Model Representation Alignment Procedure.} We select an anchor LLM space as a common coordinate system and learn ridge regression maps that project source-model representations into the anchor space. The aligned representations are then aggregated in the anchor space to form an aligned representation for downstream prediction.
  }
  \label{fig:1}
\end{figure*}

Early work primarily relied on traditional machine learning pipelines, modeling the relationship between text and market variables via manually engineered lexical, topical, and sentiment features, and achieved competitive performance across a range of settings \cite{dhar2001comparison,doyle2006extreme}. As representation learning matured, pretrained language encoders increasingly supplanted manual feature design, serving as more general and transferable sources of text representations \cite{beltagy2020longformer,huang2023finbert}. More recently, large language models (LLMs) have demonstrated stronger semantic modeling and contextual understanding, prompting growing interest in how to effectively transfer LLM capabilities to financial long-document prediction~\cite{liang2025does,zhang2025saefire,zhu2025post}. Despite these impressive gains, we identify an underappreciated source of instability that is closely tied to reliable deployment: different LLMs may encode the same long financial document through distinct inductive biases, emphasizing different evidence spans and consequently yielding heterogeneous rationales and non-identical predictions. Figure~\ref{fig:1} illustrates this phenomenon, where the same transcript activates different sets of salient phrases across models, suggesting that complementary views of the input may exist. This observation raises a practical question: can we systematically combine these heterogeneous representation spaces so that complementary evidence is converted into usable signals, rather than yielding unstable or degraded performance~\cite{lample2018word,bansal2021revisiting}. 

To address this, we propose \textbf{FinAnchor} (\underline{Fin}ancial \underline{Anchor}ed Representations), a simple and deployable framework that makes multi-encoder aggregation well-defined without fine-tuning any base model. Our approach selects one encoder as an \emph{anchor} space and learns lightweight linear alignment maps from other encoders into the anchor space using only the training split. After alignment, we aggregate the aligned representations with a fixed operator (e.g., taking the mean in the anchor space or concatenating aligned vectors) and train a small downstream readout on top. This design keeps training and inference overhead low, preserves auditability, and enables systematic reuse of multiple pretrained encoders in a unified representation space. We evaluate the proposed framework on multiple financial prediction tasks, including stock movement prediction from earnings-call transcripts and classification of central bank communications, using time-based splits to assess out-of-time generalization. Beyond aggregate performance, we provide practitioner-oriented interpretability analyses that characterize when aligned aggregation helps, using error overlap, decision-type transitions, and confidence shifts, complemented by illustrative case studies. Our contributions include:
\begin{itemize}[leftmargin=*]\setlength\itemsep{-0.3em}
  \item We introduce a lightweight multi-encoder alignment and aggregation framework that combines frozen LLM representations in a shared anchor space without fine-tuning.
  \item We conduct systematic evaluations on multiple financial prediction tasks under time-based splits, comparing against strong single-encoder and naive multi-encoder baselines.
  \item We provide interpretability analyses to identify where gains arise, including error transition patterns, confidence shifts, and case studies.
\end{itemize}

\section{Proposed Method}
\label{sec:method}

\paragraph{Problem Setup.}
We consider supervised financial prediction from long-form documents.
Each example $i$ consists of a document $t_i$ and a label $y_i$.
We encode $t_i$ using $M$ frozen LLM encoders to obtain document representations
$\mathbf{x}^{(m)}_i \in \mathbb{R}^{d_m}$ for $m \in \{1,\dots,M\}$.
To reflect realistic deployment and to evaluate out-of-time generalization,
we use a time-based split into train, validation, and test sets.
We train only lightweight components on top of the embeddings and do not fine-tune the encoders.

\paragraph{Linear Alignment to an Anchor Space.}
Raw embeddings produced by different encoders reside in heterogeneous coordinate systems,
which makes direct aggregation ill-defined.
We address this by choosing an anchor encoder $a$ and learning a linear map that projects each source
encoder $m \neq a$ into the anchor space.
Let $\mathbf{X}^{(m)} \in \mathbb{R}^{N \times d_m}$ and $\mathbf{X}^{(a)} \in \mathbb{R}^{N \times d_a}$
be the stacked embeddings on the training set.
We standardize both sides using feature-wise normalization and fit a ridge regression without intercept:
\begin{equation}
\small
\label{eq:ridge_align}
\mathbf{W}_m
=
\arg\min_{\mathbf{W}\in\mathbb{R}^{d_m \times d_a}}
\left\lVert \tilde{\mathbf{X}}^{(m)}\mathbf{W}-\tilde{\mathbf{X}}^{(a)} \right\rVert_F^2
+
\alpha \left\lVert \mathbf{W} \right\rVert_F^2,
\end{equation}
where $\tilde{\mathbf{X}}$ denotes standardized features and $\alpha$ is the ridge regularization strength.
We fit $\mathbf{W}_m$ using the training split only, and apply it to validation and test splits to obtain
aligned embeddings $\hat{\mathbf{x}}^{(m\rightarrow a)}_i = f_m(\mathbf{x}^{(m)}_i)$ in the anchor space.
To quantify linear compatibility between embedding spaces, we report the coefficient of determination
$R^2$ on the training set for each alignment, which measures the fraction of variance in the anchor
representations explained by the aligned source representations.

\paragraph{Aligned Representation Aggregation.}
After alignment, all representations share the same coordinate system, enabling simple, fixed aggregation.
In our primary setting, we compute the mean representation in the anchor space:
\begin{equation}
\label{eq:mean_agg}
\mathbf{z}_i
=
\frac{1}{M}\left(\mathbf{x}^{(a)}_i + \sum_{m\neq a} \hat{\mathbf{x}}^{(m\rightarrow a)}_i\right).
\end{equation}

\noindent In all cases, we apply a final feature standardization fitted on the training set and reused on validation
and test sets.

\paragraph{Lightweight Readout.}
We train a small classifier $g_\theta$ on top of $\mathbf{z}_i$.
For binary prediction, we use a multilayer perceptron that outputs a logit $s_i=g_\theta(\mathbf{z}_i)$
and optimize a sigmoid cross-entropy objective.
For multi-class prediction, we use a softmax classifier trained with cross-entropy and class reweighting.
Model selection and early stopping are performed on the validation set using task-appropriate metrics.
For thresholded binary decisions, we choose the operating threshold by optimizing a criterion on the
validation set and report all test metrics using the fixed selected threshold.

\begin{algorithm}[t]
\small
\DontPrintSemicolon
\SetAlgoLined
\SetKwInput{KwIn}{Input}
\SetKwInput{KwOut}{Output}
\SetKwFor{ForEach}{for each}{do}{end}

\KwIn{Anchor encoder $h_A$; source encoders $\{h_s\}_{s\in\mathcal{S}}$; training set $\mathcal{D}_{\mathrm{tr}}=\{(x_i,y_i)\}_{i=1}^n$; ridge $\alpha$}
\KwOut{Aligners $\{W_s\}$ and readout $g$}

\BlankLine
\textbf{Fit aligners (train only).}\;
\ForEach{$s\in\mathcal{S}$}{
  $
  W_s \leftarrow \arg\min_W \sum_{i=1}^n \|h_s(x_i)W - h_A(x_i)\|_2^2 + \alpha\|W\|_F^2
  $.\;
}

\BlankLine
\textbf{Aligned aggregation.}\;
$
z_{\mathrm{align}}(x)\leftarrow \frac{1}{1+|\mathcal{S}|}
\Big(h_A(x)+\sum_{s\in\mathcal{S}} h_s(x)W_s\Big)
$.\;

\BlankLine
\textbf{Readout.}\;
Train $g$ on $\{(z_{\mathrm{align}}(x_i),y_i)\}_{i=1}^n$;\;\;predict $\hat y\leftarrow g(z_{\mathrm{align}}(x))$.\;

\caption{Proposed FinAnchor.}
\label{alg:aligned_main}
\end{algorithm}

\begin{table*}[tp]
\centering
\setlength{\tabcolsep}{4pt}
\caption{Accuracy and F1 comparison across five datasets (red/blue: best/second best per column).}
\label{tab:performance}
\resizebox{1.0\textwidth}{!}{%
\begin{tabular}{@{}l
  c c
  c c
  c c
  c c
  c c@{}}
\toprule
\multirow{2}{*}{Method}
  & \multicolumn{2}{c}{Conference Call}
  & \multicolumn{2}{c}{10-Q}
  & \multicolumn{2}{c}{FNSPID Nasdaq News}
  & \multicolumn{2}{c}{Stock Movement}
  & \multicolumn{2}{c}{FOMC} \\
\cmidrule(lr){2-3} \cmidrule(lr){4-5} \cmidrule(lr){6-7} \cmidrule(lr){8-9} \cmidrule(lr){10-11}
& {Accuracy}$\uparrow$ & {F1}$\uparrow$
& {Accuracy}$\uparrow$ & {F1}$\uparrow$
& {Accuracy}$\uparrow$ & {F1}$\uparrow$
& {Accuracy}$\uparrow$ & {F1}$\uparrow$
& {Accuracy}$\uparrow$ & {F1}$\uparrow$ \\
\midrule
Zero-shot Prompting        & 0.650 & 0.676 & 0.596 & 0.628 & 0.647 & 0.669 & 0.481 & 0.301 & 0.506 & 0.508 \\
Few-shot Prompting         & 0.694 & 0.695 & 0.611 & 0.642 & 0.685 & 0.689 & 0.520 & 0.602 & 0.542 & 0.556 \\
Longformer                 & 0.744 & 0.718 & 0.721 & 0.704 & 0.758 & 0.743 & 0.530 & \cellcolor{secondbg}\textbf{0.692} & 0.636 & 0.617 \\
Hierarchical FinBERT       & \cellcolor{secondbg}\textbf{0.772} & 0.721 & 0.735 & 0.712 & 0.771 & 0.754 & 0.538 & 0.619 & 0.636 & 0.608 \\
Llama3.1-8B                & 0.758 & 0.716 & 0.722 & 0.669 & 0.767 & 0.734 & \cellcolor{secondbg}\textbf{0.578} & 0.646 & 0.607 & 0.603 \\
Qwen3-8B                   & 0.751 & 0.714 & 0.723 & 0.662 & 0.763 & 0.732 & 0.572 & 0.673 & 0.611 & 0.604 \\
Gemma2-9B                  & 0.770 & \cellcolor{secondbg}\textbf{0.731} & \cellcolor{secondbg}\textbf{0.748} & \cellcolor{secondbg}\textbf{0.718} & \cellcolor{secondbg}\textbf{0.791} & \cellcolor{secondbg}\textbf{0.764} & 0.570 & 0.665 & \cellcolor{secondbg}\textbf{0.638} & \cellcolor{secondbg}\textbf{0.640} \\
\midrule
{\textbf{FinAnchor (Ours)}} & \cellcolor{bestbg}\textbf{0.787} & \cellcolor{bestbg}\textbf{0.741}
                           & \cellcolor{bestbg}\textbf{0.755} & \cellcolor{bestbg}\textbf{0.740}
                           & \cellcolor{bestbg}\textbf{0.812} & \cellcolor{bestbg}\textbf{0.769}
                           & \cellcolor{bestbg}\textbf{0.599} & \cellcolor{bestbg}\textbf{0.693}
                           & \cellcolor{bestbg}\textbf{0.643} & \cellcolor{bestbg}\textbf{0.656} \\
\bottomrule
\end{tabular}%
}
\end{table*}

\section{Experiments}
\noindent In this section, we evaluate our FinAnchor to answer the following questions:
\textbf{RQ1:} How does the performance of FinAnchor compare to baseline models?
\textbf{RQ2:} Are the gains from the FinAnchor driven by stable, systematic signal rather than incidental noise?
\textbf{RQ3:} Does FinAnchor have the benefit of interpretability?

\subsection{Experimental Setups}
\paragraph{Tasks and Datasets.} We consider the following three finance prediction tasks.
\begin{itemize}
[leftmargin=*]\setlength\itemsep{-0.3em}
\item \emph{Earnings Surprise Prediction.}
We consider a binary text classification task that predicts whether a firm exhibits an earnings surprise from financial documents.
Inputs are long-form financial texts, and the label indicates whether the realized earnings outcome deviates from market expectations under a standard event-study style definition.
We evaluate on three commonly used corpora: {Conference Call} transcripts, {10-Q} filings, and {FNSPID Nasdaq News} articles.

\item \emph{Stock Movement Prediction.}
We consider a binary classification task that predicts next-day stock movement following an earnings call.
The input is the earnings call transcript, and the label is \{0,1\} indicating whether the stock price decreases or increases on the next trading day, using price data from Yahoo Finance.

\item \emph{FOMC Stance Classification.}
We study a three-way classification task on Federal Open Market Committee communications, predicting the monetary policy stance expressed in each document as dovish, hawkish, or neutral.
We use the FOMC Communication corpus, which aggregates FOMC meeting minutes, press conference transcripts, and speeches from 1996 to 2022, and is widely used to analyze how policy communication relates to financial market reactions.
\end{itemize}

\begin{table}[t]
\centering
\setlength{\tabcolsep}{7pt}
\begin{tabular}{lccc}
\toprule
 & \textbf{Gemma} & \textbf{Llama} & \textbf{Qwen} \\
\midrule
\textbf{Gemma} & 1.000 & 0.811 & 0.802 \\
\textbf{Llama} & 0.782 & 1.000 & 0.819 \\
\textbf{Qwen}  & 0.791 & 0.838 & 1.000 \\
\bottomrule
\end{tabular}
\caption{Error overlap (\%) between model pairs on the test set. Entry $(A,B)$ denotes $P(B\ \text{error}\mid A\ \text{error})$.}
\label{tab:error_overlap_matrix}
\end{table}

\paragraph{Comparing Baselines.} We compare with the following five families of baselines.
\begin{itemize}
[leftmargin=*]\setlength\itemsep{-0.3em}
\item \emph{LLM Zero-shot prompting.}
We apply the Gemma 3-12B-IT model in a pure zero-shot setting.
For each input, we prepend a task-specific instruction and decode greedily (temperature=0). No demonstrations or parameter updates are used.

\item \emph{LLM Few-shot prompting.}
Because Gemma 2's context window cannot hold multiple examples plus the query, we instead use Gemma 3-12B-IT with $k=4$ in-context demonstrations sampled from the training set, followed by the test input. Decoding remains greedy.

\item \emph{Longformer.} We fine-tune the Longformer-base-4096 model \citep{beltagy2020longformer}. Please refer to Section~B.3 in the Appendix for more details.

\item \emph{Hierarchical FinBERT.} We implement the Hierarchical FinBERT model from \citet{huang2023finbert} for long-document classification.

\item \emph{Single LLM Representation.}
We use the last layer hidden states from a single LLM as fixed document representations and train the same lightweight classifier as in FinAnchor, without any cross-model alignment or multi-model averaging. This baseline isolates the contribution of multi-model alignment and combination beyond simply choosing a different backbone.
\end{itemize}

\paragraph{Implementation details.}
We use Gemma2-9B-IT as the anchor model and Qwen3-8B and Llama~3.1-8B-IT as source models. For each document, we truncate inputs to at most 20{,}000 tokens and extract a document level representation from the last layer hidden states. To align multiple models representations, we learn a ridge map from each source embedding space to the anchor space on the training split only, using regularization coefficient $\alpha=10$ and no intercept, and then compute the mean of aligned representations in the anchor space. Our MLP readout  employ PyTorch with three hidden layers of size 256 and dropout of 0.5 and ReLU activations. Table~\ref{tab:performance} reports Accuracy and F1-score, and we additionally report ROC-AUC in the Appendix~\ref{appendix:experiments}.

\subsection{Comparison with Baseline Models}
We compare FinAnchor with a broad set of baselines on five datasets: Conference Call transcripts, 10-Q reports, FNSPID Nasdaq News, Stock Movement, and FOMC. Results in Table~\ref{tab:performance} yield four key insights. First, prompt-based methods lag behind learned encoders, with the largest gaps on long document tasks (Conference Call and 10-Q). Second, traditional long-document encoders remain competitive: Longformer and Hierarchical FinBERT are strong on the three text-heavy datasets, often matching or exceeding the single-LLM baseline. Third, FinAnchor delivers the best overall performance: it achieves the highest Accuracy and F1 on all five datasets, outperforming the strongest single-model LLM baselines by clear margins on Conference Call, 10-Q, and FNSPID Nasdaq News. Finally, on the more challenging Stock Movement and FOMC tasks, FinAnchor still attains the top results (0.599/0.693 and 0.643/0.656 in Accuracy/F1), indicating that our proposed FinAnchor method remains beneficial even when the signal is weaker.

\subsection{Alignment Analysis}
We investigate whether the FinAnchor’s gains reflect systematic signal rather than random fluctuations by following a simple evidence chain: (i) base representations exhibit complementary error patterns, (ii) such diversity creates room for correction that materializes as concrete decision transitions, and (iii) the changes are directionally consistent with the ground truth in terms of confidence.

\smallskip
\noindent\textbf{Error Overlaps.}
We begin with the most direct diagnostic: do different LLM representations make the same mistakes?
Table~\ref{tab:error_overlap_matrix} reports pairwise \emph{error-overlap rates} between single-model readouts.
Across model pairs, the overlap is substantially below $1$ (roughly in the $0.78$--$0.84$ range in our setting), indicating that each representation makes a non-trivial fraction of errors that the others do not.
This non-identical failure pattern is the key prerequisite for an aligned model to correct anchor errors, rather than simply reproducing them.

\begin{figure}[t]
  \centering
  \includegraphics[width=0.4\textwidth]{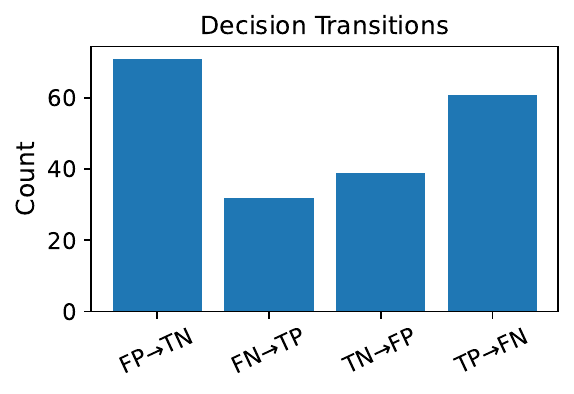}
  \caption{\textbf{Decision transitions after alignment on Stock Movement Prediction task.}
  Counts of label transitions from \textit{Gemma} to the \textit{FinAnchor} under validation-chosen thresholds. 
  The FinAnchor corrects a substantial number of false positives (FP$\rightarrow$TN), while also introducing smaller regressions.}
  \label{fig:decision-transitions}
\end{figure}

\smallskip
\noindent\textbf{Decision Transitions.}
Given this diversity, we next ask whether alignment turns it into actionable corrections.
Figure~\ref{fig:decision-transitions} summarizes how predictions move among \textsc{TP}, \textsc{FP}, \textsc{FN}, and \textsc{TN} when replacing the anchor readout (Gemma) with the aligned model (thresholds chosen on validation).
Rather than a symmetric reshuffling, the FinAnchor exhibits asymmetric shifts that correspond to concrete error corrections.
Most notably, it converts a sizable number of false positives into true negatives (\textsc{FP}$\rightarrow$\textsc{TN}: $71$), indicating fewer spurious ``up'' calls on genuinely down days.
This direction of change is particularly meaningful in financial prediction, where false positives can trigger unnecessary long positions or risk-on exposure, leading to transaction costs and avoidable downside risk, whereas false negatives often correspond to missed upside opportunities.
The prominent \textsc{FP}$\rightarrow$\textsc{TN} mass suggests that alignment yields a more risk-control-friendly behavior and provides tangible utility beyond small changes in aggregate metrics.

\begin{figure}[t]
    \centering
    \includegraphics[width=0.48\textwidth]{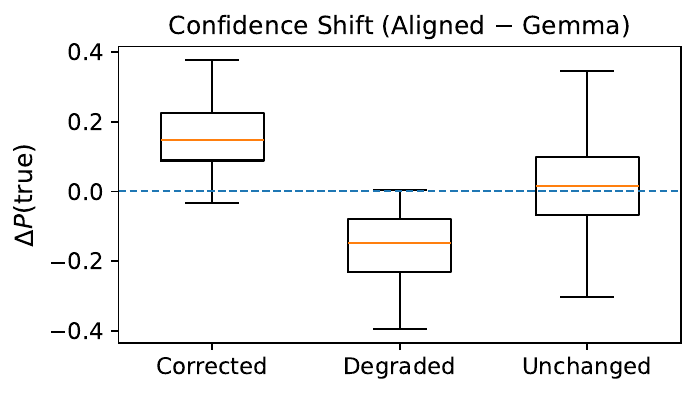}
    \caption{\textbf{FinAnchor increases confidence on corrected cases.}
    We measure the change in confidence assigned to the ground-truth label,
    $\Delta P(y)=P_{\text{Aligned}}(y)-P_{\text{Gemma}}(y)$, on the test set and stratify examples into three groups:
    \emph{Corrected} (Gemma incorrect, Aligned correct), \emph{Degraded} (Gemma correct, Aligned incorrect), and \emph{Unchanged}.
    Boxplots show the distribution of $\Delta P(y)$ within each group (median and interquartile range; whiskers denote 1.5$\times$IQR; outliers omitted).
    Positive values indicate that the FinAnchor assigns higher probability to the true label than Gemma.}
    \label{fig:delta_conf_true_boxplot}
\end{figure}

\smallskip
\noindent\textbf{Confidence Shifts.}
Decision flips alone could still arise from unstable drift around the threshold.
We therefore measure whether alignment systematically reallocates probability mass toward the ground-truth label.
For each test instance $(x,y)$, let $p_{\mathrm{G}}(y \mid x)$ and $p_{\mathrm{A}}(y \mid x)$ denote the predicted probability assigned to the true label by Gemma and the FinAnchor, respectively.
We define the confidence shift:
$
\Delta p_{\mathrm{true}}(x) \;=\; p_{\mathrm{A}}(y \mid x) \;-\; p_{\mathrm{G}}(y \mid x).
$
If the FinAnchor improves by extracting stable signal, we expect $\Delta p_{\mathrm{true}}$ to be positive on corrected cases (Gemma $\times \rightarrow$ Aligned $\checkmark$), and close to zero on unchanged cases.
Figure~\ref{fig:delta_conf_true_boxplot} matches this expectation: corrected examples exhibit a clear positive shift, whereas degraded/unchanged groups concentrate much closer to $0$.
This suggests that the FinAnchor’s improvements are driven by directionally consistent confidence increases on the true label, not arbitrary perturbations.

\begin{table}[t]
\centering
\begin{tabular}{l r}
\toprule
Source $\rightarrow$ Anchor & Train $R^2$ \\
\midrule
Llama-3.1-8B $\rightarrow$ Gemma-2-9B & 0.97 \\
Qwen3-8B $\rightarrow$ Gemma-2-9B & 0.98 \\
\bottomrule
\end{tabular}
\vspace{-2mm}
\caption{Linear representation alignment quality measured by train-set $R^2$ on the anchor space.}
\label{tab:align_r2}
\end{table}

\begin{table*}[t]
\centering
\small
\setlength{\tabcolsep}{4pt}
\renewcommand{\arraystretch}{1.3}
\begin{tabular}{r p{0.58\textwidth} c c c}
\toprule
\textbf{Idx} & \textbf{Sentence (full)} & \textbf{Gemma} & \textbf{Llama} & \textbf{FinAnchor} \\
\midrule
205 &
First quarter, as the industry is having more \kw{promotional activity} and we were price, we saw fairly significant increase in demand and also a corresponding reduction in \kw{gross margin}.
& $\downarrow\,-0.040$ & $\downarrow\,-0.380$ & $\downarrow\,-0.807$ \\

62 &
The heightened \kw{promotional environment} that we experienced in the first quarter moderated over the last several months, resulting in comparable \kw{gross margin} rates versus last year.
& $\downarrow\,-0.031$ & $\downarrow\,-0.365$ & $\downarrow\,-0.795$ \\

160 &
We continue to see that business \kw{growing} in the \kw{mid- to high-20s} range, and so that's the continuing funding of that business.
& $\downarrow\,-0.033$ & $\downarrow\,-0.358$ & $\downarrow\,-0.780$ \\

156 &
I wanted to follow up on the \kw{growth in inventory} this quarter and specifically, the \kw{planned growth} in full-line.
& $\downarrow\,-0.025$ & $\downarrow\,-0.363$ & $\downarrow\,-0.772$ \\

61 &
Our \kw{gross profit rate} decreased 7 basis points from last year primarily due to higher \kw{occupancy expenses} related to Rack's growth.
& $\downarrow\,-0.023$ & $\downarrow\,-0.351$ & $\downarrow\,-0.770$ \\
\bottomrule
\end{tabular}
\vspace{2pt}
\caption{Sentence-level occlusion on a false-positive example for Gemma-only that is corrected by the FinAnchor (ground truth: \textit{down}). For each sentence $i$, we report $\Delta z_i = z_{\text{base}} - z_{-i}$, where $z$ is the \textit{up} logit and $z_{-i}$ is the logit after removing sentence $i$. Arrows indicate direction: $\uparrow$ supports \textit{up} ($\Delta z>0$), while $\downarrow$ supports \textit{down} ($\Delta z<0$). Highlighted spans mark salient financial cues.}
\label{tab:case-occlusion}
\end{table*}

\subsection{Interpretability Analysis}
We further examine what changes in the decision process after representation alignment, beyond aggregate performance.
Our analysis proceeds in two steps.
First, we quantify how well each source representation can be mapped into the anchor space.
Second, we inspect a representative corrected example to reveal which textual evidence is amplified by the FinAnchor.

\paragraph{Linear alignment quality.}
Table~\ref{tab:align_r2} reports the coefficient of determination ($R^2$) for the ridge regression map that aligns each source representation to the anchor (Gemma) embedding space.
Concretely, for a source model $s$, we learn a linear map $T_s$ on the training set by minimizing
\begin{equation}
  \label{eq:align_ridge}
  T_s \;=\; \arg\min_T \sum_{i \in \mathcal{D}_{\text{tr}}} \left\lVert T\,x_i^{(s)} - x_i^{(\text{G})} \right\rVert_2^2 \;+\; \lambda \lVert T \rVert_2^2,
\end{equation}
where $x_i^{(s)}$ is the representation from source $s$ for example $i$ and $x_i^{(\text{G})}$ is the corresponding Gemma representation.
We compute $R^2$ on the training set:
\begin{equation}
\small
  \label{eq:align_r2}
  R^2 \;=\; 1 - \frac{\sum_{i \in \mathcal{D}_{\text{tr}}} \left\lVert x_i^{(\text{G})} - T_s(x_i^{(s)}) \right\rVert_2^2}
  {\sum_{i \in \mathcal{D}_{\text{tr}}} \left\lVert x_i^{(\text{G})} - \bar{x}^{(\text{G})} \right\rVert_2^2},
\end{equation}
with $\bar{x}^{(\text{G})}$ denoting the mean anchor representation.
High $R^2$ indicates that a large fraction of the anchor-space variance is explainable via a simple linear re-parameterization of the source features.
This provides evidence that different LLM representations share a compatible geometric structure in this domain, and that alignment plausibly converts different coordinate systems into a common semantic frame, making subsequent fusion meaningful rather than an arbitrary numeric averaging.

\paragraph{Case study: evidence reweighting under alignment.}
We next illustrate how alignment changes the evidence that drives a prediction.
Table~\ref{tab:case-occlusion} presents a false-positive example under Gemma-only that is corrected by the FinAnchor (Stock Movement; transcript id $683961$; ground truth: \emph{down}).
To measure sentence-level influence, we perform occlusion by removing a sentence $i$ and recomputing the \emph{up-class logit} $z$.
For each model, we report
$\Delta z_i \;=\; z_{\text{base}} - z_{-i},$
where $z_{\text{base}}$ is the up-logit on the full transcript and $z_{-i}$ is the up-logit after removing sentence $i$.
A positive $\Delta z_i$ indicates that sentence $i$ supports the \emph{up} prediction (removing it decreases the up-logit), while a negative $\Delta z_i$ indicates that sentence $i$ supports \emph{down} (removing it increases the up-logit).

The key pattern in Table~\ref{tab:case-occlusion} is that the FinAnchor assigns substantially larger magnitude $|\Delta z_i|$ to financially salient, bearish cues that are only weakly reflected in Gemma-only.
Specifically, multiple sentences describing margin compression, cost pressures, and inventory buildup receive small influence under Gemma-only, yet are strongly weighted by the FinAnchor.
These cues are consistent with a conservative interpretation of near-term fundamentals and provide a plausible mechanism for correcting the false positive: by amplifying shared bearish evidence across views in a common coordinate system, the FinAnchor reduces reliance on weaker or non-diagnostic portions of the transcript and shifts the decision toward the true \emph{down} label.
Overall, the case study supports that alignment changes decisions via structured evidence reweighting, rather than random perturbations of confidence.

\section{Conclusions}
We present a simple alignment approach for combining heterogeneous LLM representations: we learn linear mappings into an anchor space, aggregate the aligned views, and train a lightweight readout for downstream financial prediction tasks.
Across five tasks, the FinAnchor improves over single-view baselines while remaining easy to implement.
Our analyses suggest the improvements are not noise: different views make non-identical errors, and alignment turns this diversity into actionable corrections, particularly by reducing false positives that correspond to spurious buy signals. Our interpretability analyses further indicate that the gains arise from aligning complementary semantic evidence across models, rather than from random perturbations.

\clearpage
\section*{Limitations}
Our experiments and analyses focus on finance-oriented prediction settings, using a fixed set of backbone models under a unified evaluation protocol. As a result, the generality of our findings should be further validated on a broader range of tasks, languages, and model families. In addition, our approach combines frozen representations with an alignment and aggregation module, which keeps training and deployment simple but introduces additional implementation choices and hyperparameters (e.g., alignment strength and aggregation strategy). A more systematic sensitivity study and an engineering-oriented cost analysis would help strengthen reproducibility and practical adoption.

\bibliography{custom}

\begin{thebibliography}{26}
\providecommand{\natexlab}[1]{#1}

\bibitem[{Bansal et~al.(2021)Bansal, Nakkiran, and Barak}]{bansal2021revisiting}
Yamini Bansal, Preetum Nakkiran, and Boaz Barak. 2021.
\newblock Revisiting model stitching to compare neural representations.
\newblock \emph{Advances in neural information processing systems}, 34:225--236.

\bibitem[{Beltagy et~al.(2020)Beltagy, Peters, and Cohan}]{beltagy2020longformer}
Iz~Beltagy, Matthew~E. Peters, and Arman Cohan. 2020.
\newblock Longformer: The long-document transformer.
\newblock \emph{arXiv preprint arXiv:2004.05150}.

\bibitem[{Brown et~al.(2004)Brown, Hillegeist, and Lo}]{brown2004conference}
Stephen Brown, Stephen~A Hillegeist, and Kin Lo. 2004.
\newblock Conference calls and information asymmetry.
\newblock \emph{Journal of Accounting and Economics}, 37(3):343--366.

\bibitem[{Bushee et~al.(2018)Bushee, Gow, and Taylor}]{bushee2018linguistic}
Brian~J Bushee, Ian~D Gow, and Daniel~J Taylor. 2018.
\newblock Linguistic complexity in firm disclosures: Obfuscation or information?
\newblock \emph{Journal of Accounting Research}, 56(1):85--121.

\bibitem[{Dhar and Chou(2001)}]{dhar2001comparison}
Vasant Dhar and Dashin Chou. 2001.
\newblock A comparison of nonlinear methods for predicting earnings surprises and returns.
\newblock \emph{IEEE Transactions on Neural networks}, 12(4):907--921.

\bibitem[{Dong et~al.(2024)Dong, Fan, and Peng}]{dong2024fnspid}
Zihan Dong, Xinyu Fan, and Zhiyuan Peng. 2024.
\newblock \href {https://arxiv.org/abs/2402.06698} {Fnspid: A comprehensive financial news dataset in time series}.
\newblock \emph{Preprint}, arXiv:2402.06698.

\bibitem[{Doyle et~al.(2006)Doyle, Lundholm, and Soliman}]{doyle2006extreme}
Jeffrey~T Doyle, Russell~J Lundholm, and Mark~T Soliman. 2006.
\newblock The extreme future stock returns following i/b/e/s earnings surprises.
\newblock \emph{Journal of Accounting Research}, 44(5):849--887.

\bibitem[{{Eli Bartov}(1992)}]{eli_bartov_patterns_1992}
{Eli Bartov}. 1992.
\newblock \href {http://www.jstor.org/stable/247981} {Patterns in {Unexpected} {Earnings} as an {Explanation} for {Post}-{Announcement} {Drift}}.
\newblock \emph{The Accounting Review}, 67(3):610--622.
\newblock Publisher: American Accounting Association.

\bibitem[{Huang et~al.(2023)Huang, Wang, and Yang}]{huang2023finbert}
Allen~H Huang, Hui Wang, and Yi~Yang. 2023.
\newblock Finbert: A large language model for extracting information from financial text.
\newblock \emph{Contemporary Accounting Research}, 40(2):806--841.

\bibitem[{Koval et~al.(2023)Koval, Andrews, and Yan}]{koval2023forecasting}
Ross Koval, Nicholas Andrews, and Xifeng Yan. 2023.
\newblock Forecasting earnings surprises from conference call transcripts.
\newblock In \emph{Findings of the Association for Computational Linguistics: ACL 2023}, pages 8197--8209.

\bibitem[{Lample et~al.(2018)Lample, Conneau, Ranzato, Denoyer, and Jégou}]{lample2018word}
Guillaume Lample, Alexis Conneau, Marc'Aurelio Ranzato, Ludovic Denoyer, and Hervé Jégou. 2018.
\newblock \href {https://openreview.net/forum?id=H196sainb} {Word translation without parallel data}.
\newblock In \emph{International Conference on Learning Representations}.

\bibitem[{Larcker and Zakolyukina(2012)}]{larcker2012detecting}
David~F Larcker and Anastasia~A Zakolyukina. 2012.
\newblock Detecting deceptive discussions in conference calls.
\newblock \emph{Journal of Accounting Research}, 50(2):495--540.

\bibitem[{Latane and Jones(1979)}]{latane1979standardized}
Henry~A Latane and Charles~P Jones. 1979.
\newblock Standardized unexpected earnings--1971-77.
\newblock \emph{The Journal of Finance}, 34(3):717--724.

\bibitem[{Liang and Carrasco~Kind(2025)}]{liang2025does}
Qingwen Liang and Matias Carrasco~Kind. 2025.
\newblock How does managers' willingness to disclose affect analysts' earning forecasts-a measurement by llms.
\newblock \emph{Available at SSRN 5199752}.

\bibitem[{Loughran and McDonald(2011)}]{loughran2011liability}
Tim Loughran and Bill McDonald. 2011.
\newblock Liability or asset? the role of language in financial reports.
\newblock \emph{Journal of Accounting Research}, 49(1):135--178.

\bibitem[{Luo et~al.(2022)Luo, Ravina, Sammon, and Viceira}]{luo_retail_2022}
Cheng Luo, Enrichetta Ravina, Marco Sammon, and Luis~M. Viceira. 2022.
\newblock \href {https://doi.org/10.2139/ssrn.3544949} {Retail investors’ contrarian behavior around news, attention, and the momentum effect}.
\newblock Technical report, Social Science Research Network.
\newblock Posted: 5 Apr 2020; Last revised: 26 May 2022.

\bibitem[{Qin and Yang(2019)}]{qin2019what}
Yu~Qin and Yi~Yang. 2019.
\newblock What you say and how you say it matters: Predicting stock volatility using verbal and vocal cues.
\newblock In \emph{Proceedings of the 57th Annual Meeting of the Association for Computational Linguistics}, pages 390--401. Association for Computational Linguistics.

\bibitem[{Sang and Bao(2022)}]{sang2022predicting}
Yunxin Sang and Yang Bao. 2022.
\newblock Predicting corporate risk by jointly modeling company networks and dialogues in earnings conference calls.
\newblock \emph{arXiv preprint arXiv:2206.06174}.

\bibitem[{Sawhney et~al.(2020)Sawhney, Mathur, Mangal, Khanna, Shah, and Zimmermann}]{sawhney2020multimodal}
Ramit Sawhney, Puneet Mathur, Ayush Mangal, Piyush Khanna, Rajiv~Ratn Shah, and Roger Zimmermann. 2020.
\newblock Multimodal multi-task financial risk forecasting.
\newblock In \emph{Proceedings of the 28th ACM International Conference on Multimedia}, pages 456--465. ACM.

\bibitem[{Shah et~al.(2023)Shah, Paturi, and Chava}]{shah-etal-2023-trillion}
Agam Shah, Suvan Paturi, and Sudheer Chava. 2023.
\newblock \href {https://doi.org/10.18653/v1/2023.acl-long.368} {Trillion dollar words: A new financial dataset, task {\&} market analysis}.
\newblock In \emph{Proceedings of the 61st Annual Meeting of the Association for Computational Linguistics (Volume 1: Long Papers)}, pages 6664--6679, Toronto, Canada. Association for Computational Linguistics.

\bibitem[{Si et~al.(2013)Si, Mukherjee, Liu, Li, Li, and Deng}]{si2013exploiting}
Jianfeng Si, Arjun Mukherjee, Bing Liu, Qing Li, Huayi Li, and Xiaotie Deng. 2013.
\newblock Exploiting topic based twitter sentiment for stock prediction.
\newblock In \emph{Proceedings of the 51st Annual Meeting of the Association for Computational Linguistics (Volume 2: Short Papers)}, pages 24--29.

\bibitem[{Skinner and Sloan(2002)}]{skinner2002earnings}
Douglas~J Skinner and Richard~G Sloan. 2002.
\newblock Earnings surprises, growth expectations, and stock returns or don't let an earnings torpedo sink your portfolio.
\newblock \emph{Review of accounting studies}, 7(2):289--312.

\bibitem[{Yang et~al.(2020)Yang, Zhang, Li, Bendersky, and Najork}]{yang2020beyond}
Liu Yang, Mingyang Zhang, Cheng Li, Michael Bendersky, and Marc Najork. 2020.
\newblock Beyond 512 tokens: Siamese multi-depth transformer-based hierarchical encoder for long-form document matching.
\newblock In \emph{Proceedings of the 29th ACM International Conference on Information \& Knowledge Management (CIKM)}, pages 1725--1734. ACM.

\bibitem[{Zaheer et~al.(2020)Zaheer, Guruganesh, Dubey, Ainslie, Alberti, Ontanon, Pham, Ravula, Wang, Yang et~al.}]{zaheer2020big}
Manzil Zaheer, Guru Guruganesh, Kumar~Avinava Dubey, Joshua Ainslie, Chris Alberti, Santiago Ontanon, Philip Pham, Anirudh Ravula, Qifan Wang, Li~Yang, and 1 others. 2020.
\newblock Big bird: Transformers for longer sequences.
\newblock \emph{Advances in neural information processing systems}, 33:17283--17297.

\bibitem[{Zhang et~al.(2025)Zhang, Liu, Zhang, He, and Du}]{zhang2025saefire}
Huopu Zhang, Yanguang Liu, Miao Zhang, Zirui He, and Mengnan Du. 2025.
\newblock \href {https://arxiv.org/abs/2505.14420} {Sae-fire: Enhancing earnings surprise predictions through sparse autoencoder feature selection}.
\newblock \emph{Preprint}, arXiv:2505.14420.

\bibitem[{Zhu et~al.(2025)Zhu, Liu, and Sheng}]{zhu2025post}
Yu~Zhu, Xiao Liu, and Olivia R~Liu Sheng. 2025.
\newblock Post-earnings-announcement drift prediction: Leveraging postevent investor responses with multitask learning.
\newblock \emph{Information Systems Research}.

\end{thebibliography}

\clearpage
\appendix

\section{Related Work}
Financial prediction tasks provide rich forward-looking information, driving substantial research interest. Early studies, such as \citep{larcker2012detecting} and \citep{bushee2018linguistic}, focus on sentiment extraction and its impact on analyst forecasts. Recent advances apply deep learning to directly predict earnings surprises \citep{koval2023forecasting}. 
Multimodal models further enhance prediction accuracy by integrating text, audio, and network structures \citep{qin2019what, sawhney2020multimodal, sang2022predicting, yang2020beyond}. \citet{huang2023finbert} develop FinBERT, a domain-specific language model that significantly improves financial text understanding.

\section{More Experimental Details}
\subsection{Experimental Settings}
\noindent\textbf{Supervised Learning Task.}
We frame the prediction of the direction of the next earnings surprise (\( ES \)) as a supervised learning task using only the textual content of the most recent financial documents. Following \citet{latane1979standardized}, we measure \( ES \) using the Standardized Unexpected Earnings (SUE), defined as the difference between reported EPS and the analyst consensus estimate, scaled by the standard deviation of analyst forecasts. 
The consensus estimate is computed as the mean of the latest valid analyst forecasts issued within one month after the earnings call, allowing analysts to update their expectations based on the call content and recent financial disclosures. This setup provides a forward-looking measure of market expectations and presents a more realistic yet challenging prediction task. The average time horizon between the input document and the target earnings event is approximately three months, further underscoring the difficulty of the task.
\begin{equation}
\small
ES = \frac{RepEPS - \text{Avg}(EstEPS)}{\text{Std}(EstEPS)}
\end{equation}
\begin{equation}
\small
y =
\begin{cases}
0, & ES \leq -\delta \\
1, & ES \geq \delta
\end{cases}
\end{equation}
We transform the continuous earnings surprise (\( ES \)) into a binary classification task by assigning a label of \( +1 \) for \( ES > \delta \) and \( 0 \) for \( ES < -\delta \), where \( \delta = 0.50 \). This choice follows previous studies on standardized unexpected earnings \citep{eli_bartov_patterns_1992} and on price momentum \citep{luo_retail_2022} , where earnings surprise are considered large when \(|SUE| \ge 0.5\).This threshold balances sample size and event significance. Documents with immaterial surprises (i.e., \( ES \in [-0.50, 0.50] \)) are excluded, as these near-zero values often elicit weak market responses and may reflect earnings management. We evaluate performance using accuracy and train models with binary cross-entropy loss.
Although \( ES \) is a continuous variable, market reactions are typically binary, responding more to the direction of the surprise than to its magnitude. Our focus is therefore on material surprises that are more likely to influence investor behavior and pricing.

\subsection{Datasets}\label{appendix:sec:datasets}
\textbf{Conference Call.} We manually collected English-language earnings call transcripts for the largest publicly traded U.S. firms from the Seeking Alpha website. The dataset consists of English-language transcripts from quarterly earnings conference calls of publicly traded U.S. companies.The transcripts were annotated with the
names of the speakers (executives and analysts) and content, offering rich, high-frequency textual and contextual data for studying corporate communication, sentiment, and strategic disclosures. To focus on liquid and economically significant firms, we restrict the sample to companies with market capitalizations exceeding \$1 billion and average daily trading volumes above \$50 million. Our sample included 9,324 earnings conference calls, between 2012 and 2014.

\noindent
\textbf{10\,-Q Report.} The second dataset comprises the \textit{Management Discussion and Analysis} (MDA) sections from firms’ quarterly (10Q) filings with the U.S. Securities and Exchange Commission (SEC). The MDA provides management’s narrative perspective on past performance, future outlook, and key business risks, offering rich forward-looking information. We focus on this section because it reflects direct managerial communication to shareholders. All texts are preprocessed to remove boilerplate language, tables, and footnotes to retain core narrative content.

\begin{figure*}[t]
  \centering
  \includegraphics[width=\textwidth]{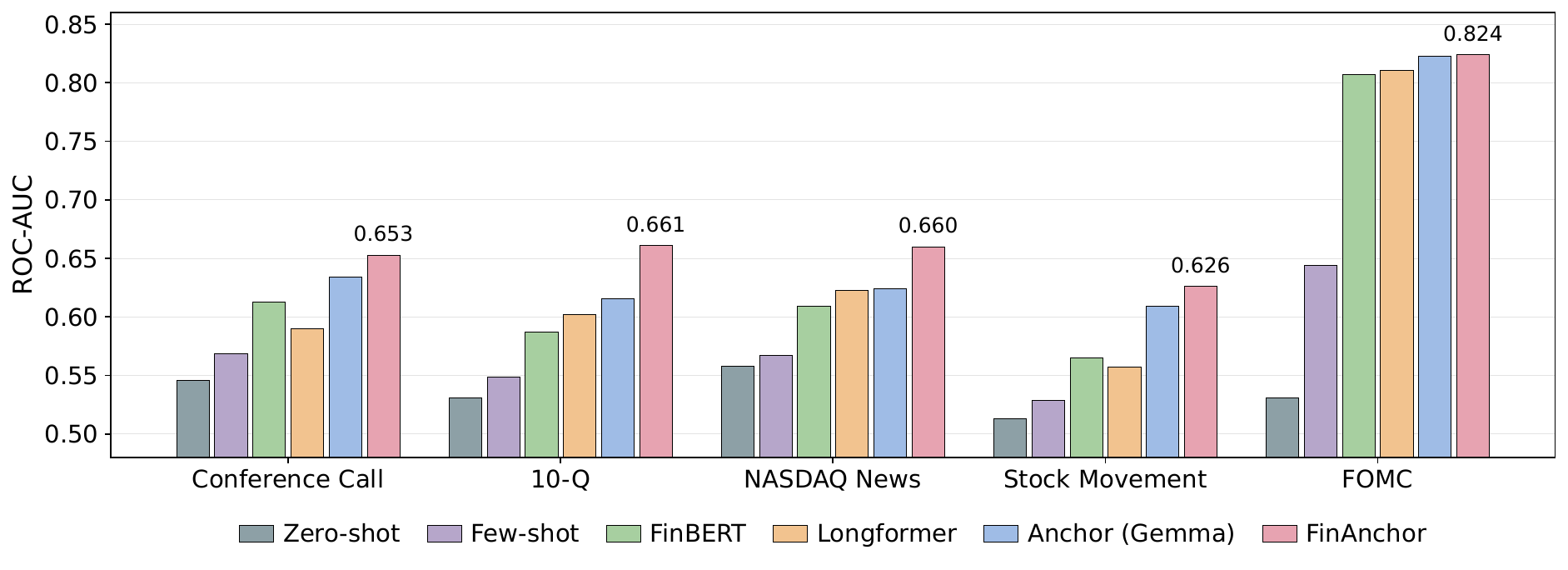}
  \caption{\textbf{ROC-AUC comparison across five datasets.} We report ROC-AUC for six methods (Zero-shot, Few-shot, FinBERT, Longformer, Anchor (Gemma), and FinAnchor), grouped by dataset.}
  \label{fig:auc}
\end{figure*}

\noindent
\textbf{FNSPID Nasdaq News.} The third dataset is \textit{FNSPID Financial News and Stock Price Integration Dataset}, a large-scale time-series dataset that integrates textual and market data \cite{dong2024fnspid}. It includes 29.7~million stock price observations and 15.7~million financial news records for 4,775 S\&P~500 firms from 1999 to 2023, collected from four major news websites. It provides summaries of financial news records generated using various methods as well. Given the scale of this corpus, processing full articles for a single firm over a quarter would result in inputs orders of magnitude longer than those in the other two datasets. Therefore, we use the dataset’s LSA summaries: for each firm–quarter we sort summaries chronologically and concatenate them into a single document formatted as a numbered list covering the entire quarter preceding the next EPS announcement. This retains broad topical coverage while keeping length manageable and comparable across datasets. FNSPID stands out for its scale, diversity, and sentiment signals derived from news, which have been shown to enhance transformer-based stock return predictions. It also provides a reproducible updating pipeline with code and documentation, making it a valuable resource for financial research.

\noindent\textbf{Financial Data.} We obtained Reported Earnings Per Share (EPS) and analyst consensus EPS forecasts from the IBES database. For each earnings call, we align realized EPS values with analyst consensus forecasts, allowing for the computation of earnings surprises. The combination of rich textual data with structured earnings performance metrics makes this dataset especially suitable for tasks such as managerial sentiment analysis and earnings surprise prediction.

\noindent
\textbf{FOMC (Trillion Dollar Words).}
The FOMC dataset is built from Federal Open Market Committee (FOMC) communications in~\cite{shah-etal-2023-trillion}, covering three types of official texts: meeting minutes, Chair press conference transcripts, and speeches by Federal Reserve officials. It spans 1996--2022 (with more complete press conference coverage starting in 2011). The authors clean the raw documents, segment them into sentences, and apply a keyword-driven filtering and sampling procedure to select policy-relevant sentences for annotation. The core task is a three-way classification of policy stance (hawkish/dovish/neutral), corresponding to tightening, easing, or maintaining the current stance; the dataset also includes metadata (e.g., document type and time) to support comparisons across sources. The dataset is released on Hugging Face as \texttt{gtfintechlab/fomc\_communication}, providing sentence-level text (\texttt{sentence}), year (\texttt{year}), and label (\texttt{label}), along with the official \texttt{train}/\texttt{test} splits (about 2.48k instances in total, with 496 in \texttt{test}). We directly follow the Hugging Face release and its splits in our experiments.

\section{More Experimental Results}
\subsection{ROC-AUC Results}
We provide ROC-AUC results for all methods and datasets in Figure~\ref{fig:auc} (computed under the same evaluation protocol as the main experiments). For FOMC, which is a multi-class task, we report macro-averaged ROC-AUC (one-vs-rest).

\subsection{Concatenation Without Alignment.}
As an ablation, we evaluate a baseline that directly concatenates representations from different LLMs without performing alignment, followed by the same prediction head. We find this approach is consistently weaker than alignment-based aggregation and even underperforms the best single-LLM baseline. Here, ``Concatenation'' denotes feature-level concatenation: for each input document, we extract the representation vector from each LLM and concatenate these vectors along the feature dimension to form a single document representation, which is then fed into the same prediction head (without any alignment). We report this lightweight ablation only on FOMC, since direct concatenation without alignment is not well-motivated and we do not expect it to generalize across tasks.

\begin{table}[t]
\centering
\small
\setlength{\tabcolsep}{6pt}
\begin{tabular}{lccc}
\toprule
\textbf{Model / Setting} & \textbf{F1} & \textbf{Acc} & \textbf{ROC-AUC} \\
\midrule
Source (Llama)  & 0.603 & 0.607 & 0.807 \\
Source (Qwen)   & 0.604 & 0.611 & 0.811 \\
Anchor (Gemma)  & 0.640 & 0.638 & 0.823 \\
Concatenation (no alignment) & 0.619 & 0.613 & 0.805 \\
\textbf{FinAnchor} & \textbf{0.656} & \textbf{0.643} & \textbf{0.824} \\
\bottomrule
\end{tabular}
\caption{FOMC results for a lightweight ablation study (F1/Acc/ROC-AUC).}
\label{tab:fomc_ablation}
\end{table}

\label{appendix:experiments}

\end{document}